# YOLOv5 vs. YOLOv8 in Marine Fisheries: Balancing Class Detection and Instance Count

Mahmudul Islam Masum
*School of Computing and Information Sciences*
Florida International University
Miami, USA
mmasu004@fiu.edu

Arif Sarwat
*Department of Electrical and Computer Engineering*
Florida International University
Miami, USA
asarwat@fiu.edu

Hugo Riggs
*Department of Electrical and Computer Engineering*
Florida International University
Miami, USA
hrigg002@fiu.edu

Alicia Boymelgreen
*Department of Mechanical and Materials Engineering*
Florida International University
Miami, USA
aboymelg@fiu.edu

Preyojon Dey
*Department of Mechanical and Materials Engineering*
Florida International University
Miami, USA
pdey004@fiu.edu

*Abstract*— **This paper presents a comparative study of object detection using YOLOv5 and YOLOv8 for three distinct classes: artemia, cyst, and excrement. In this comparative study, we analyze the performance of these models in terms of accuracy, precision, recall, etc. where YOLOv5 often performed better in detecting Artemia and cysts with excellent precision and accuracy. However, when it came to detecting excrement, YOLOv5 faced notable challenges and limitations. This suggests that YOLOv8 offers greater versatility and adaptability in detection tasks while YOLOv5 may struggle in difficult situations and may need further fine-tuning or specialized training to enhance its performance. The results show insights into the suitability of YOLOv5 and YOLOv8 for detecting objects in challenging marine environments, with implications for applications such as ecological research.**

*Keywords—fisheries, artemia, YOLOv5, YOLOv8, nanoparticle*

## I. INTRODUCTION

Object detection technology became a powerful tool in fisheries management and research that enabled automated identification and tracking of objects. This study focuses on the comparative analysis of two cutting-edge object detection models, YOLOv5 and YOLOv8. We analyzed their performance in detecting three important objects within aquatic environments: live *Artemia* (small zooplankton), cysts, and excrement. Accurate detection of these objects is essential for various fisheries-related tasks as it helps us understand ecosystem dynamics, assess fish health, and optimize aquaculture operations.

We designed an experimental setup and procedure to conduct the study. The setup was crucial for the free movement of *Artemia* (brine shrimp) within a controlled environment.

*Artemia* are widely used as live feed in marine fisheries due to their high nutrition content. Their high nutritional density and ease of culture make it a valuable resource for feeding marine larvae with small mouth gapes. This contribution to the development and growth of marine organisms makes *Artemia* essential in aquaculture. The small size, rapid movement, and variable orientations of *Artemia* pose a unique challenge for object detection which makes them an ideal test subject for evaluating the capabilities of YOLOv5 and YOLOv8 in dynamic fisheries environments. The data for our dataset were obtained during *Artemia* cyst hatching in various nanoparticle-saturated saltwater samples. The uptake was captured with microscopic Image sampling as the nanoparticles are fluorescent.

Capturing nanoparticle uptakes in challenging environments faces a common set of challenges due to the limitations of microscopy. One of the well-known issues in microscopy is that the images captured in these challenging scenarios are often out of focus and blurry. These challenges are particularly common when studying nanoparticle uptake in microfluidic systems where small and dynamic subjects are dealt with. However, the microscopy approach remains advantageous for small-scale microfluidic studies.

In previous studies [1], researchers have shown that different components like *Artemia*, cysts, and excrements within a microfluidic environment offer valuable insights into the metabolic processes of these organisms. This microfluidic platform provides a high level of control over the microenvironment and enables direct observation of morphological changes. It helps researchers monitor and differentiate between various stages of the hatching process of *Artemia*. By subjecting *Artemia* to different temperatures and salinities, significant alterations in the duration of hatching stages, metabolic rates, and hatchability are observed. For example, it is found that higher temperatures and moderate salinity significantly enhance the metabolic resumption of dormant Artemia cysts. This demonstrates the critical role that environmental factors play in influencing metabolic activity.

Object detection models, such as YOLO (You Only Look Once) play a significant role in advancing the field of computer vision with their real-time capabilities and competitive accuracy. While YOLOv5 is the more optimized version, YOLOv8 further refines the architecture, aiming to enhance both accuracy and speed. By conducting a comparative study using this experimental setup, we aim to provide practical insights into how these models perform and their applicability to fisheries management.

## II. LITERATURE REVIEW

In recent years, the field of object detection particularly in applications related to fisheries and underwater environments has witnessed remarkable progress.

A study [2] introduced a novel multitask model that combined the YOLOv5 architecture with a semantic segmentation head for real-time fish detection and segmentation. Experiments conducted on the golden crucian carp dataset showcase an impressive precision of 95.4% in object detection and semantic segmentation accuracy of 98.5%. The model also exhibits competitive performance on the PASCAL VOC 2007 dataset, achieving object detection precision of 73.8% and semantic segmentation accuracy of 84.3%. The model worked with a remarkable speed by achieving processing rates of 116.6 FPS and 120 FPS on an RTX3060. The YOLOv5's capabilities were enhanced due to the addition of a segmentation head. The study followed a precise methodology, including model validation, selection, ablation experiments, and optimization, and compared its results with other models.

Researchers [3] addressed the challenge of detecting fish in dense groups and small underwater targets with the introduction of an enhanced CME-YOLOv5 network. This algorithm achieves a mean average precision (mAP@0.50) of 94.9%, surpassing the baseline YOLOv5 by 4.4 percentage points. Notably, it excels in detecting densely spaced fish and small targets, making it a promising tool for fishery resource investigation. The algorithm effectively mitigates issues related to missing detections in dense fish schools and enhances accuracy for small target fish with limited pixel and feature information. Moreover, it demonstrates proficiency in detecting small objects and effectively handling occluded and highly overlapping objects, outperforming YOLOv5 by detecting 49 additional objects and achieving a 24.6% increase in the detection ratio.

One study that centers on improving underwater object detection by evaluating various models, including EfficientDet, YOLOv5, YOLOv8, and Detectron2, on the "Brackish-Dataset" of annotated underwater images captured in Limfjorden water [4]. The research project aims to evaluate the efficiency of these models in terms of accuracy and inference time while proposing modifications to enhance EfficientDet's performance. The study conducts a thorough comparison of multiple object detection models, revealing that EfficientDet had the lead with a mean Average Precision (mAP) of 98.56%, followed closely by YOLOv8 with 98.20% mAP. YOLOv5 secures a mAP of 97.6%. Notably, the modified EfficientDet, incorporating a modified BiSkFPN mechanism and adversarial noise handling, achieves an outstanding mAP of 98.63%, with adversarial learning also elevating YOLOv5's accuracy to 98.04% mAP. The research also provided class activation map-based explanations (CAM) for the two models to promote explainability in black box models.

In terms of addressing inefficient, and potentially damaging manual sorting methods, another study [5] introduced an automated, non-contact sorting approach for crayfish using an enhanced YOLOv5 algorithm. The proposed algorithm attains a mean Average Precision (mAP) of 98.8% while reducing image processing time to a mere 2 ms. It was significant in crayfish sorting, offering an accurate, efficient, and fast alternative. The reduction in image processing time significantly enhances the overall speed of the algorithm, optimizing crayfish sorting operations.

A modified version of YOLOv5, called UTD-Yolov5 was designed and introduced in another study [6] to identify the Crown of Thorns Starfish (COTS) in underwater images. This algorithm is crucial for complex underwater operations. UTD-Yolov5 incorporates several modifications to the YOLOv5 network architecture, including a two-stage cascaded CSP backbone, a visual channel attention mechanism module, and a random anchor box similarity calculation method. Additionally, optimization methods such as Weighted Box Fusion (WBF) and iterative refinement are proposed to enhance network efficiency. Experiments conducted on the CSIRO dataset revealed that UTD-Yolov5 achieves an average accuracy of 78.54% which is a significant improvement over the baseline.

## III. METHODOLOGY OF THE STUDY

In this experimental study, researchers conducted imaging of *Artemia* nauplii in various saltwater samples saturated with nanoparticles. The process involved capturing images of the nauplii under a microscope. The *Artemia* nauplii were freely moving in some samples and others were selected for fixed sample capture.

*A. Dataset*

The dataset employed in this study consisted of 1000 images, of which 800 were allocated for training, and 200 for validation purposes. To optimize computational efficiency and save computational resources, all images underwent a format transformation. Originally in TIF format, these images were converted to PNG format. This conversion served a dual purpose: it significantly reduced image size while enhancing computational efficiency.

By switching to the PNG format, the dataset became clearly visible and more manageable without compromising the quality of the images. Furthermore, to facilitate model training with YOLO and to optimize computational efficiency, all images were resized from their original dimensions of 2048x2044 to a standardized size of 640x640 pixels. This resizing ensured uniformity across the dataset while reducing computational overhead.

When the images were resized to a standardized size of 640x640 pixels, the nauplii in the images became smaller in terms of pixel dimensions compared to their original, larger-sized images which can affect the measured speed of the nauplii (calculated as pixels per frame). This means that the same physical movement of a nauplius would correspond to fewer pixels in the resized images than in the original images. As a result, the calculated speed in pixels per frame might appear artificially high. Therefore, it's essential to consider the specific scale factor and the potential difference in pixel values relative to real-world instances and distances.

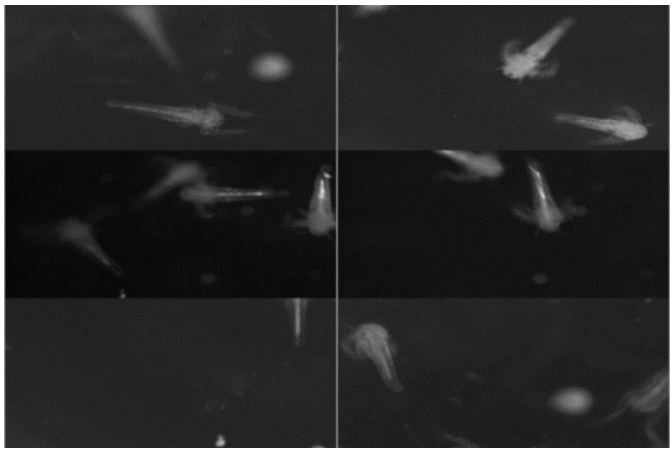

Figure 1. Sample of images from the dataset

The selection of these 1000 images was performed manually from a larger collection of 2250 images. This manual curation process was undertaken to ensure better clarity and accurate positioning of objects within the images. By carefully handpicking the images, the dataset was refined to emphasize high-quality and representative samples, enhancing the reliability of the object detection models in capturing and recognizing *Artemia*, Cyst, and Excrement.

### B. Class-instances

The training dataset consisted of a total of 1,716 annotated class instances as follows: Artemia holds the majority with 1,368 instances, making up a substantial 79.7% of the total instances. Cyst is represented by 297 instances which is 17.3% of the training set. Lastly, excrement has 51 annotated instances, approximately 3% of the training dataset.

In the validation set, we have a total of 435 class instances. Here, the class distribution follows a similar trend. Artemia is the most dominant class with 355 instances, approximately 81.6% of the validation set. Cysts are represented by 66 instances, contributing 15.2% to the validation dataset. Excrement appeared only 14 times, just 3.2% of the total validation instances. The similar allocation of instances between the training and validation datasets allowed us to comprehensively evaluate the models' performance on two datasets while maintaining a proportional representation of the three classes.

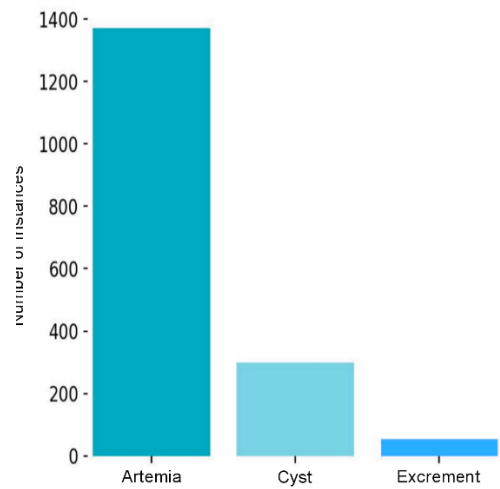

Figure 2. Number of class-instances

The total dataset combining both training and validation sets has the following class distribution: Artemia stands out as the dominant class with 1,723 instances or roughly 80.1% of the entire dataset. This majority presence of the Artemia instances reflects their ecological significance in marine environments. Cysts make up 36.3% of the dataset, with a total of 363 instances. The least common class is just 6.5% of the dataset, with 65 instances. Recognizing this class distribution is essential for our research, as it provides the basis for evaluating the object detection models' performance.

### C. Structural Similarity

The samples of this dataset were derived from three distinct concentrations: 50mg, 100mg, and a controlled concentration. Specifically, it includes 400 images each from the 50mg and 100mg concentrations. 200 more image samples are collected from the controlled concentration. This distribution ensures a balanced representation of various concentration levels. The importance of the Structural Similarity Index (SSIM) in this context cannot be overstated. SSIM is a metric used to measure the similarity between two images. In our study, it plays a crucial role in assessing the similarities between images in the dataset. This assessment is vital for several reasons because by evaluating the SSIM, we can ensure that the images across different concentrations maintain a consistent quality which is crucial for accurate analysis and comparison.

The SSIM also helps in determining how static or dynamic the samples appear in terms of structure and appearance. This is essential when studying the impact of different concentrations on the subjects. A higher SSIM would indicate more structural similarity and less dynamic change, whereas a lower SSIM might suggest significant alterations due to concentration differences. It helps in identifying any distinct structural changes that occur due to varying concentrations.

Another value that is being measured is MSE (Mean Squared Error). It is a metric to measure the average squared difference between the pixel values of the two compared images. MSE value of 0 indicates a perfect similarity when an image is being compared with itself. The MSE value is greater than 0 for different images where higher values indicating greater differences between the images.

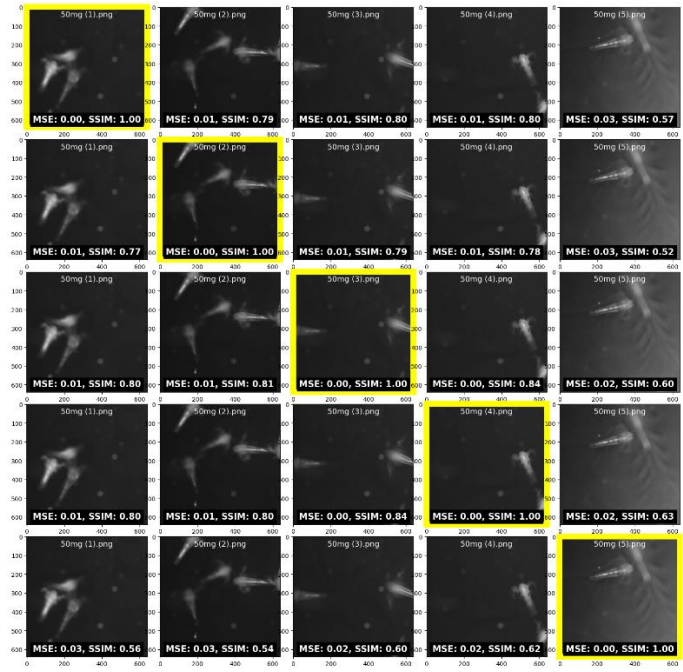

Figure 3. SSIM and MSE values of some sample of 50mg concentration

The MSE values from figure 3 represents the average of the squares of the differences between the pixel intensities of two images. As mentioned earlier, an MSE of 0 indicates no difference between the compared images (perfect similarity). For instance, when '50mg (1).png' is compared with itself, the MSE is 0.00 which indicates a perfect match. At the same time, the SSIM values measure the similarity between two images in terms of luminance, contrast, and structure. SSIM values range from -1 (no similarity) to 1 (perfect similarity). For example, '50mg (1).png' compared with itself has an SSIM of 1.00, which is expected since it's the same image. There are five sample from each concentration to measure and evaluate these values. Images that have a yellow border are the reference images for that row. It means all the MSE and SSIM values displayed at the bottom of other images in that row are calculated in comparison to the reference image with the yellow border. For example, in the second row, '50mg (2).png' has a yellow border, indicating it is the reference image for comparisons in that row. '50mg (1).png' has comparisons with MSE values ranging from 0.00 to 0.03 and SSIM values from 0.52 to 1.00, while '50mg (2).png' has comparisons with MSE values ranging from 0.00 to 0.03 and SSIM values from 0.54 to 1.00. '50mg (3).png' has comparisons with MSE values ranging from 0.00 to 0.03 and SSIM values from 0.56 to 1.00, while '50mg (4).png' has comparisons with MSE values ranging from 0.00 to 0.02 and SSIM values from 0.60 to 1.00. The last image, '50mg (5).png' has MSE values ranging from 0.00 to 0.03 and SSIM values from 0.52 to 1.00.

Considering the low MSE values and the high SSIM values, it can be said that the images are very similar to each other in 50mg concentration. This high degree of similarity indicates that there are minimal changes in texture, luminance, and contrast between the images.

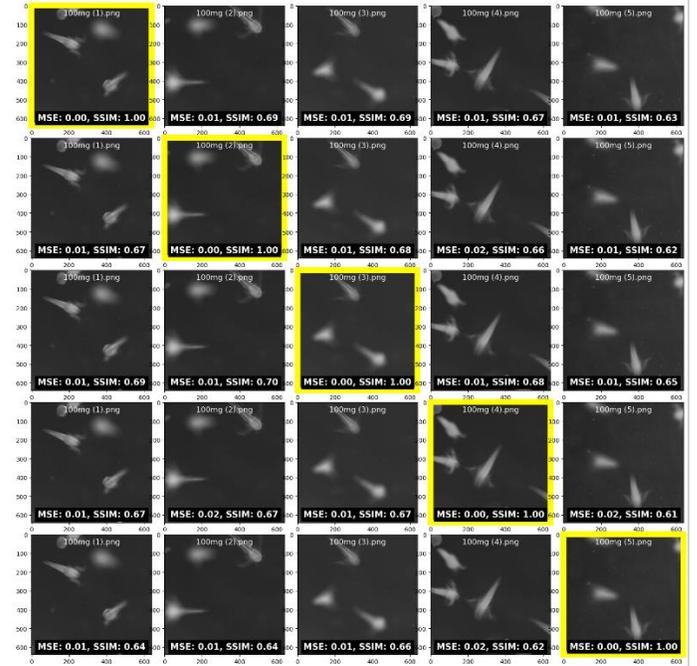

Figure 4. SSIM and MSE values of some sample of 100mg concentration

In figure 4, the MSE values for the comparisons are again low, measured ranging from 0.01 to 0.02 which indicates that the pixel intensity differences between the images are minor while the SSIM values are mostly in the range of 0.61 to 0.70, with some reaching 1.00 (when comparing an image with itself). These values are a bit lower than in the previous set which indicates a slight decrease in structural similarity, but they still reflect a high degree of similarity overall. '100mg (1).png' has comparisons with MSE values ranging from 0.00 to 0.02 and SSIM values from 0.61 to 1.00, while '100mg (2).png' has comparisons with MSE values ranging from 0.00 to 0.02 and SSIM values from 0.62 to 1.00. '100mg (3).png' has comparisons with MSE values ranging from 0.00 to 0.01 and SSIM values from 0.64 to 1.00, while '100mg (4).png' has comparisons with MSE values ranging from 0.00 to 0.02 and SSIM values from 0.66 to 1.00. '100mg (5).png' has comparisons with MSE values ranging from 0.00 to 0.02 and SSIM values from 0.61 to 1.00.

The patterns indicate that when comparing different images, there is a slight decrease in SSIM values compared to the previous set (50mg). It suggests that these images are a bit less similar to each other than the images in the 50mg set. However, the similarities are still high, which suggests that the differences between the images might still subtle.

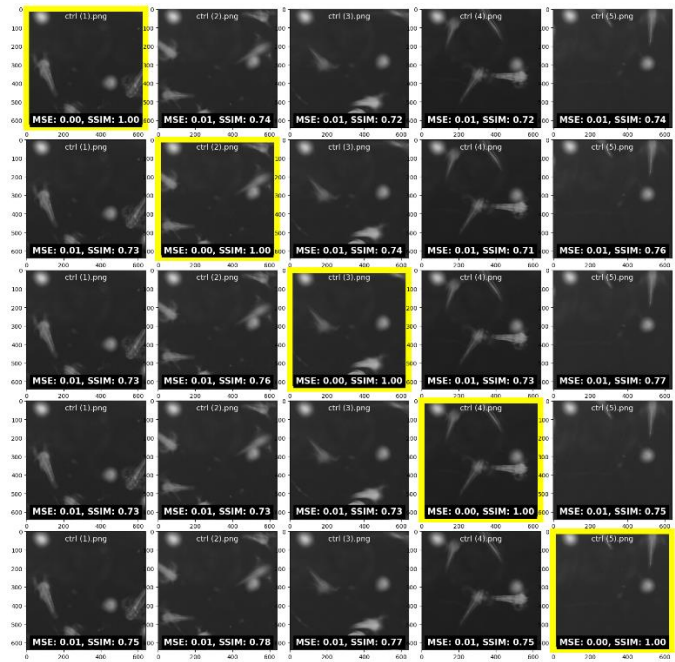

Figure 5. SSIM and MSE values of some sample of controlled concentration

In figure 5, 'ctrl (1).png' is compared to others with MSE values ranging from 0.00 to 0.01 and SSIM values from 0.72 to 1.00, while 'ctrl (2).png' shows MSE values from 0.00 to 0.01 and SSIM values from 0.71 to 1.00 when compared to others. 'ctrl (3).png' has MSE values from 0.00 to 0.01 and SSIM values from 0.72 to 1.00 in its comparisons while 'ctrl (4).png' is compared with others showing MSE values from 0.00 to 0.01 and SSIM values from 0.71 to 1.00. Finally, 'ctrl (5).png' has MSE values from 0.00 to 0.01 and SSIM values from 0.72 to 1.00 in comparisons.

The controlled concentration images exhibit consistently low Mean Squared Error (MSE) values and high Structural Similarity Index (SSIM) values across comparisons. These metrics indicate a high degree of similarity among the images within the controlled set. It also indicates that there are minimal variations in terms of pixel intensity, texture, and structure. The high SSIM values and low MSE figures observed in the controlled concentration comparisons reinforce the uniformity and stability of these samples by providing a solid foundation for the integrity of the experimental design and subsequent data analysis. The consistency in these key image quality metrics facilitates a more accurate and clear understanding of the impact of concentration levels on the subjects under study.

*D. Training*

To train the object detection models, we adopted a structured training procedure. The dataset, comprising 1000 images with 800 for training and 200 for validation, was prepared as previously described. For this comparative study, we employed the power of the YOLO (You Only Look Once) object detection framework; YOLOv5 and YOLOv8. The training was conducted specifically using YOLOv5s (YOLO version 5-small) and YOLOv8s (YOLO version 8-small) pre-trained model architectures.

The training process encompassed 10 epochs. Each epoch iterated over the entire training dataset eight times (batch size of 8). This precise training process allowed the models to learn and adapt to the features and variations within the dataset. During the training process, the models optimized their internal parameters by minimizing a loss function through stochastic gradient descent (SGD) optimization that allowed the models to learn and adapt to the distinctive characteristics of the objects of interest within the dataset.

By progressively upgrading from YOLOv5 to YOLOv8, we explored the advancements in model architectures and their impact on object detection performance. The 10 training epochs and a batch size of 8 secured a balance between model convergence and computational efficiency. It also ensured that the models could effectively capture and classify *Artemia*, Cyst, and Excrement within the images.

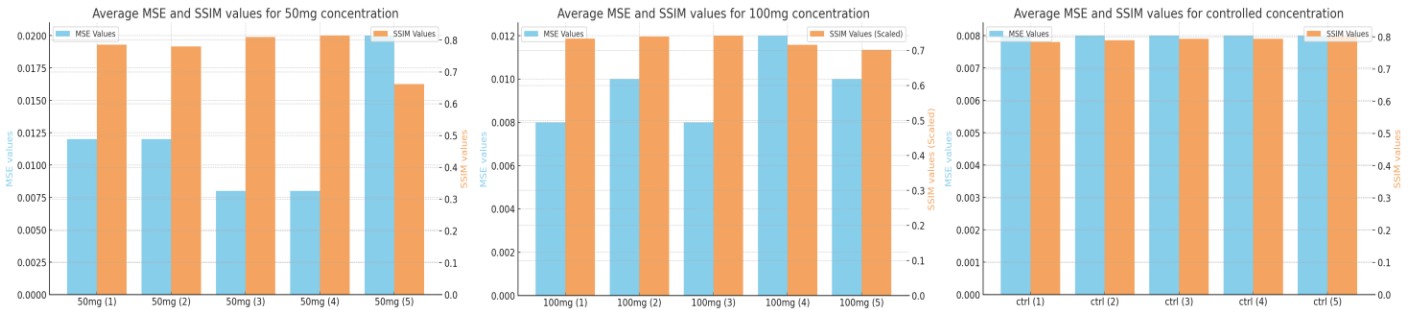

Figure 6. Average SSIM and MSE value from 50mg, 100mg, and controlled concentration

## IV. RESULT AND DISCUSSION

For object detection, we employed the YOLOv5 model and YOLOv8 model by utilizing YOLOv5s and YOLOv8s pre-trained models respectively. We primarily relied on precision, recall, F-score, and mean Average Precision (mAP) metrics to assess object detection accuracy.

Precision emphasizes on the model's ability to make accurate positive predictions, aiming to minimize false positives. Recall, on the other hand, focuses on how effectively the model detects positive instances within the dataset. These metrics are calculated using the following formulas, where "Pos" represents Positive, "Neg" represents Negative, "T" stands for True, and "F" denotes False:

$$Precision = \frac{True\ positve}{True\ Positive + False\ Negative}$$

$$Recall = \frac{T\_Positve}{T\_Positive + F\_Negative}$$

Accuracy serves as an additional metric to gauge classification performance, considering both true and false predictions.

$$Accuracy = \frac{T\ Pos + T\ Neg}{T\ Pos + T\ Neg + F\ Pos + F\ Neg}$$

$$F1 = 2 \cdot \frac{precision \cdot recall}{precision + recall}$$

The F1 Confidence Curve illustrates how sensitive the model is to the threshold used for classifying a positive instance. It also helps in identifying the threshold that provides the best balance between precision and recall. It also shows how robust the model is across different levels of confidence at the same time. It is vital in applications where the cost of false positives and false negatives varies. The F1 Confidence Curve can provide insights beyond what a single F1 score might reveal, especially in cases where models have similar F1 scores but different precision-recall balances.

Furthermore, we evaluate the overall object detection performance using the mean Average Precision (mAP). This metric represents the average of the Average Precision (AP) calculated for all the classes being detected.

### A. YOLOv5 training outcome

The training of YOLOv5 on the curated dataset delivered highly encouraging results, particularly noteworthy due to the challenging conditions presented by fisheries images. The microscopic study approach is vital for capturing detailed images in challenging aquatic environments, where images often exhibit issues, such as being out of focus and blurry. The environment becomes especially challenging when dealing with microscopic subjects like nanoparticles and Artemia nauplii. This microscopic imaging approach is essential because it allows researchers to obtain highly detailed images, which is valuable. In a study [7] to identify, classify, and separate *Larvae* using an aspiration pipette or water stream, researchers also used a microscopic approach to capture images. Despite the complexities of microscopy, YOLOv5 demonstrated its robustness and adaptability to precisely detect and analyze objects within these images.

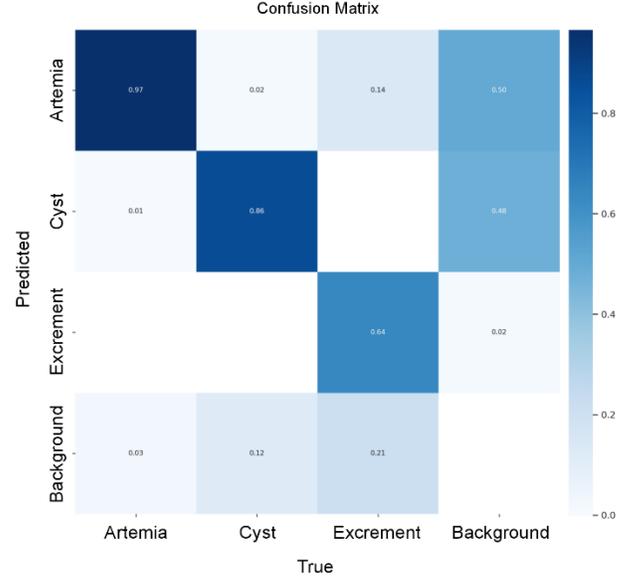

Figure 7. YOLOv5 Confusion Matrix

The confusion matrix of YOLOv5 shows that the pre-trained model had a success rate of 0.97 for Artemia, 0.86 for Cyst, and 0.64 for Excrement. The false positives were negligible for Artemia and Cyst with a rate of 0.01 and 0.02 respectively.

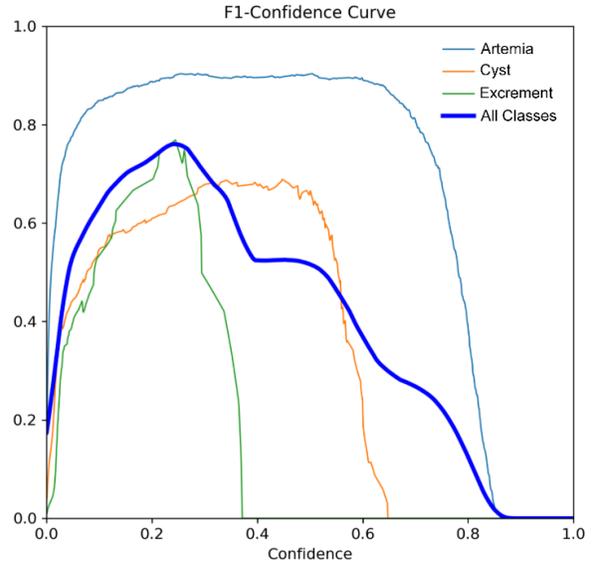

Figure 8. YOLOv5 F1 - Confidence curve

The F1 - Confidence serves as a dynamic representation of the model's precision-recall trade-off for different confidence thresholds. At a threshold of 0.241, the model consistently achieved a score of 0.76, which is a balanced measure of precision and recall. This finding signifies that the model demonstrated a remarkable ability to maintain both high precision, minimizing false positives, and high recall, effectively capturing positive instances within the dataset.

In the Precision-Recall curve, an average precision (AP) of 0.766 indicates that the model effectively minimized false positives while capturing the majority of positive instances within the dataset. The model consistently achieved a precision-recall balance at a confidence threshold of 0.5 that yielded an average precision (AP) of 0.766 across all classes.

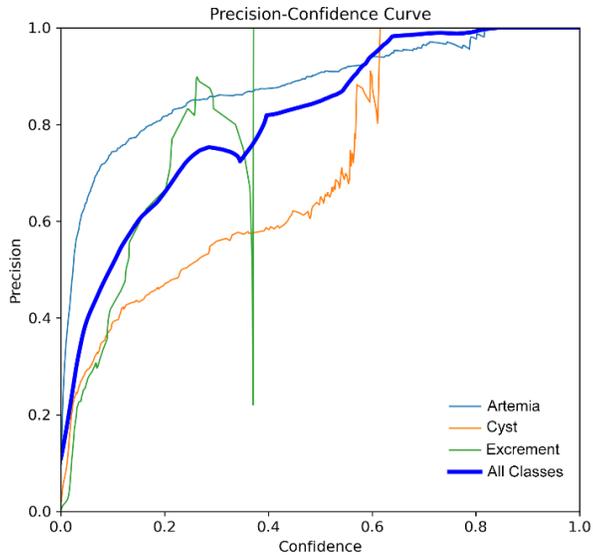

Figure 9. YOLOv5 Precision - confidence curve

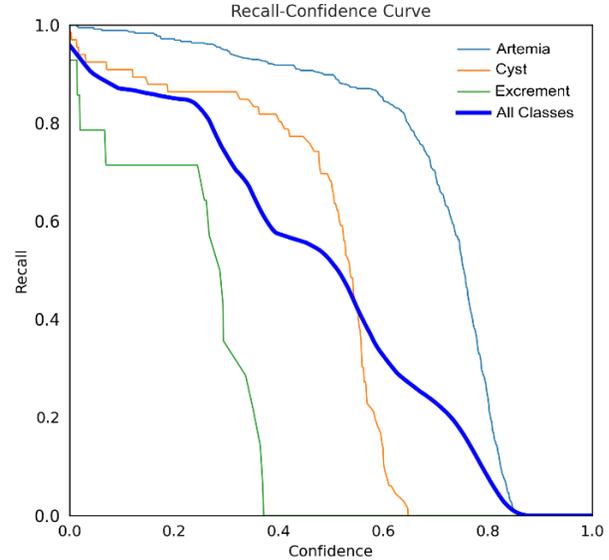

Figure 11. YOLOv5 Recall - confidence curve

The precision-confidence curve is a crucial visualization that showcases the model's precision at different confidence thresholds. A precision score of 1.00 signifies that the model made no false-positive predictions at the confidence threshold of 0.843. In other words, when the model identified an object with a confidence score exceeding 0.843, it was unequivocally accurate in its predictions.

As for recall confidence, at an extremely low confidence threshold of 0.000, the model consistently achieved an outstanding precision-recall balance, with a precision score of 0.96 across all classes.

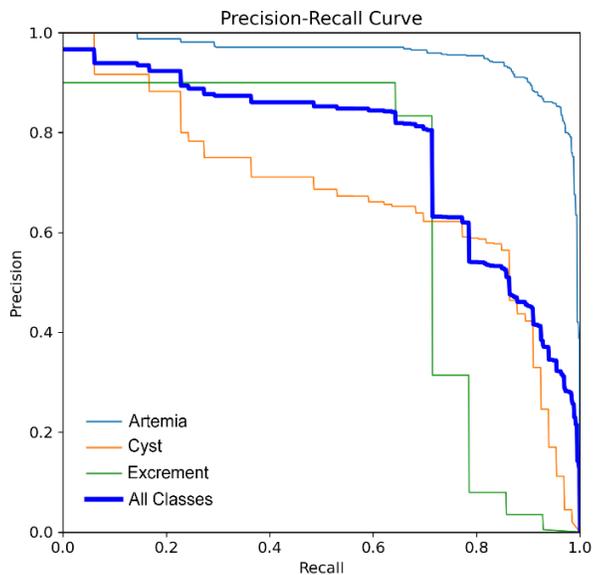

Figure 10. YOLOv5 Precision - Recall curve

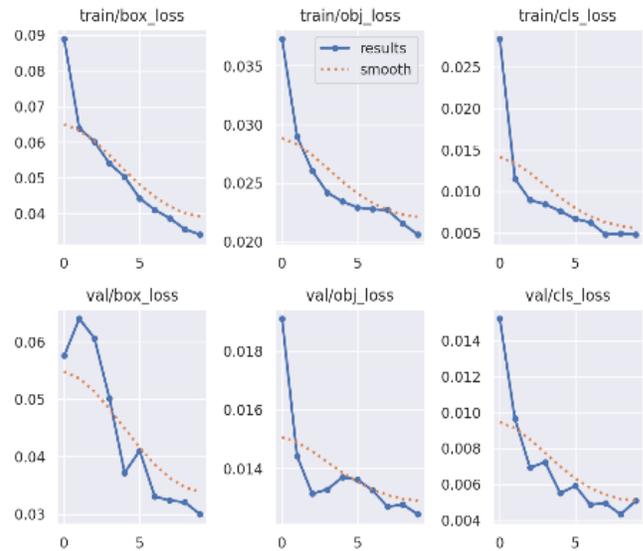

Figure 12. Loss of Boxes, Objects, and Classes in YOLOv5

The understanding of the rate of losses during both the training and validation phases is crucial for evaluating the performance of object detection models. During the initial stages of training, the loss of boxes started at approximately 0.09, indicating a reasonable starting point. As training progressed through the 10 epochs, there was a gradual and consistent improvement that resulted in a decrease in box loss. Around the 5th epoch, the rate of box loss reached a stable level which reflected the model's ability to efficiently predict bounding boxes for objects. In contrast, the validation loss curve represents a less smooth trajectory compared to the training curve. It started at approximately 0.06 and then rose to approximately 0.07, and then gradually decreased to around 0.04. After the 5th epoch, the validation curve exhibited stability and achieved a commendable box loss rate of 0.03.

The training curve showcased an initial object loss of approximately 0.035, which stabilized after just three epochs, ultimately reaching a final loss rate of 0.020. on the other hand, the validation curve displayed a more inconsistent pattern but ultimately achieved a superior loss rate of less than 0.014 at the end of training.

The class loss during training initially had a rate of over 0.025. However, after a mere 2-3 epochs, the loss rate dropped to less than 0.010 with an excellent final rate of 0.005. In the validation phase, the class loss was initially slightly above 0.014, accompanied by a few spikes in the curve. The final loss rate was around 0.005, signifying that the model excelled in inaccurate class prediction during both the training and validation phases

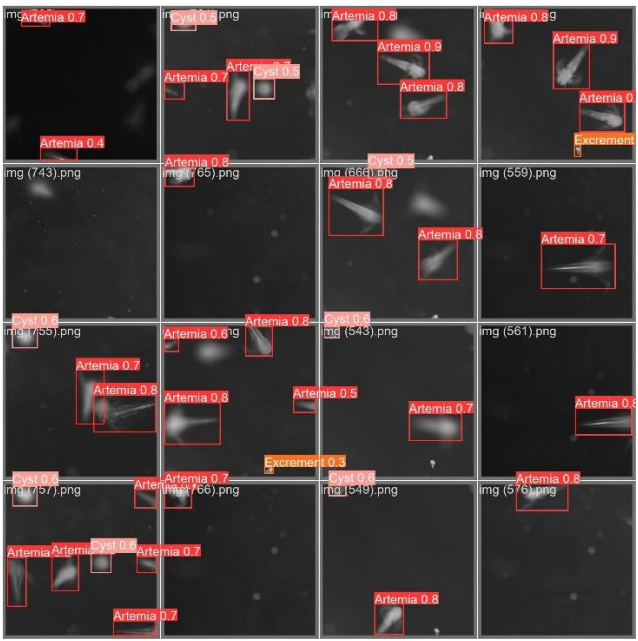

Figure 13. YOLOv5 pre-trained model training outputs

## B. *YOLOv8 training outcome*

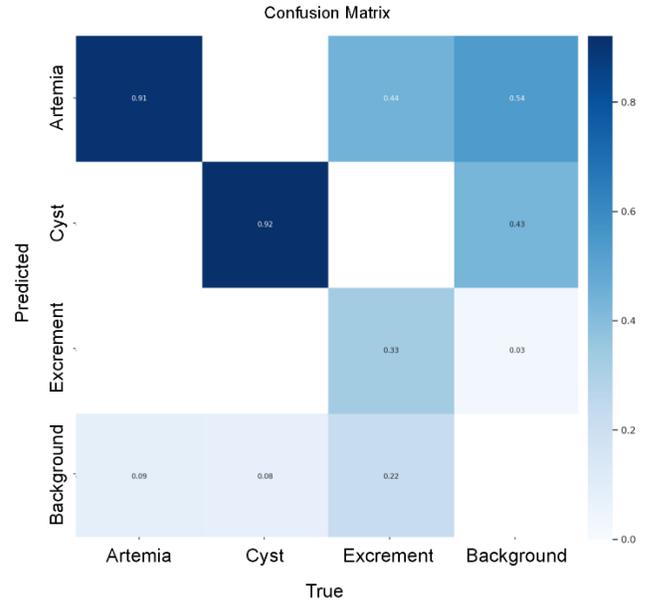

Figure 14. YOLOv8 Confusion Matrix

The confusion matrix of YOLOv8 shows that the pre-trained model had a success rate of 0.91 for Artemia, 0.92 for Cyst with 0 false positives, and 0.33 for Excrement.

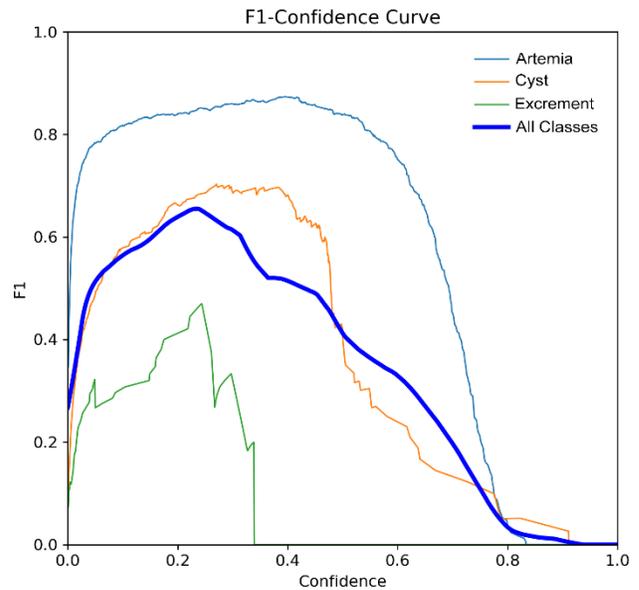

Figure 15. YOLOv8 F1 - Confidence curve

The F1 - Confidence curve for YOLOv8 illustrates the model's precision-recall trade-off across various confidence thresholds. At a threshold of 0.234, the model consistently achieved an F1 score of 0.66.

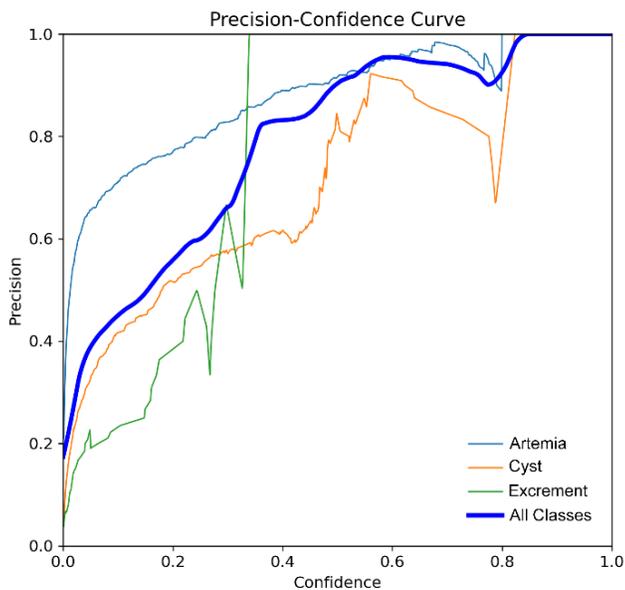

Figure 16. YOLOv8 Precision - Confidence curve

For YOLOv8, at a confidence threshold of 0.849, the YOLOv8 model consistently achieved a perfect precision score of 1.00 for all classes.

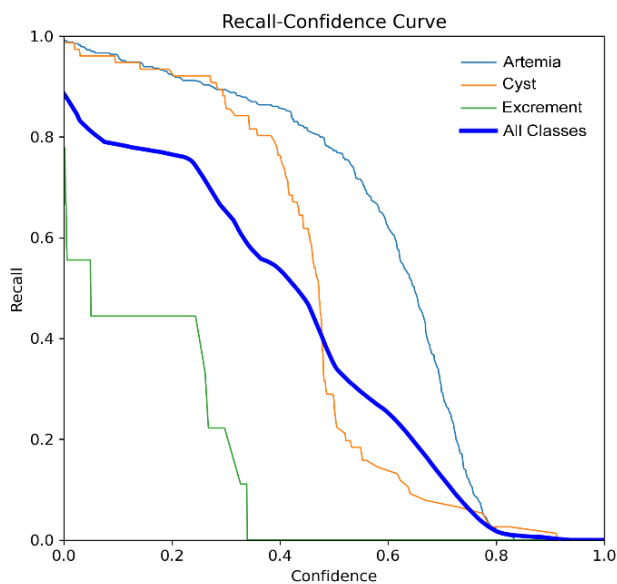

Figure 18. YOLOv8 Recall - Confidence curve

The YOLOv8 model consistently achieved an outstanding precision-recall balance with a precision score of 0.89 across all classes.

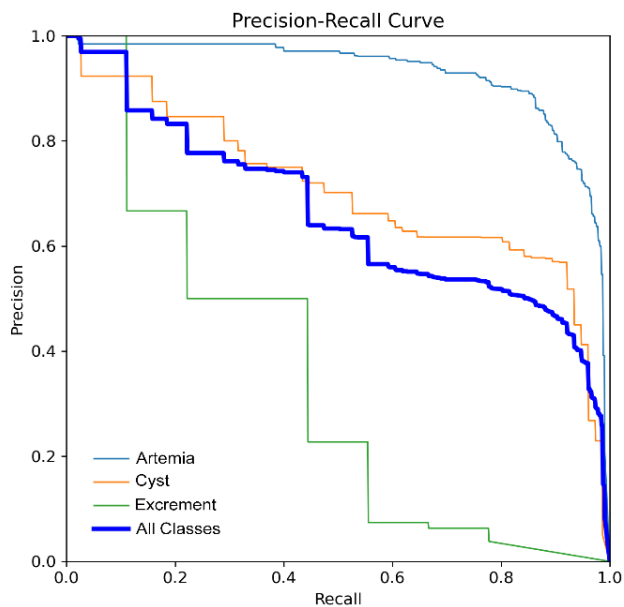

Figure 17. YOLOv8 Precision - Recall curve

At a moderate confidence threshold with mAP@0.5, the model consistently achieved a balanced precision-recall performance, resulting in an average precision (AP) of 0.658 across all classes. The precision-recall curve is an essential visualization that assesses how well the model balances precision (minimizing false positives) and recall (capturing positive instances) at varying confidence thresholds. The model maintained a balance between precision and recall with mAP@0.5.

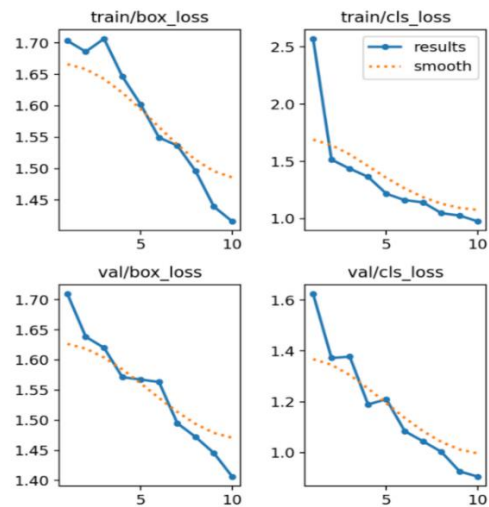

Figure 19. Loss of Boxes and Classes in YOLOv8

In training, the rate of box loss was initially at 1.70 before exhibiting a consistent downward trajectory. After 10 training epochs, the final box loss rate was at less than 1.45. The validation phase also demonstrated a notable reduction in box loss by reaching an even more impressive rate of 1.40 after 10 epochs. The class loss during training followed a similar pattern. The model initially struggled with a class loss of approximately 2.5 which indicated challenges in correctly classifying objects. However, the model made progress by reducing the class loss to a rate of 1.0 after 10 epochs. On the other hand, the model's class loss exhibited a slightly different trajectory during validation. It started at a value of 1.6 before having some spikes in the curve. However, the curve gradually smoothed out beyond the 5th epoch, reaching a final rate of approximately 0.05.

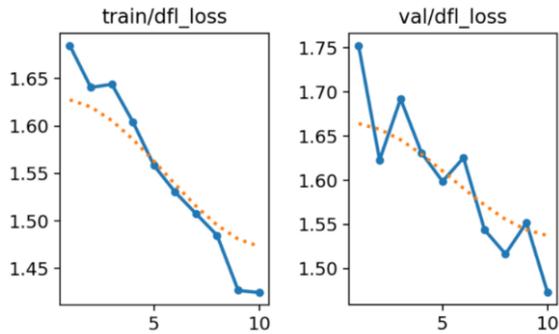

Figure 20. Distributional focal loss in YOLOv8

During the training process, the Distributional focal loss (DFL) starts at an initial value of over 1.65. However, as training progresses, the loss exhibits a consistent downward trend, ultimately converging to a final rate of less than 1.45.
In the validation phase, the DFL was initially at 1.75, indicating some initial challenges. After the first two epochs, it demonstrates an encouraging decrease, reaching nearly 1.60. However, there is again an increase in the loss, peaking at 1.70. With some spikes in the validation curve, the Distributional focal loss becomes more stable over time, with a final loss rate of less than 1.50.

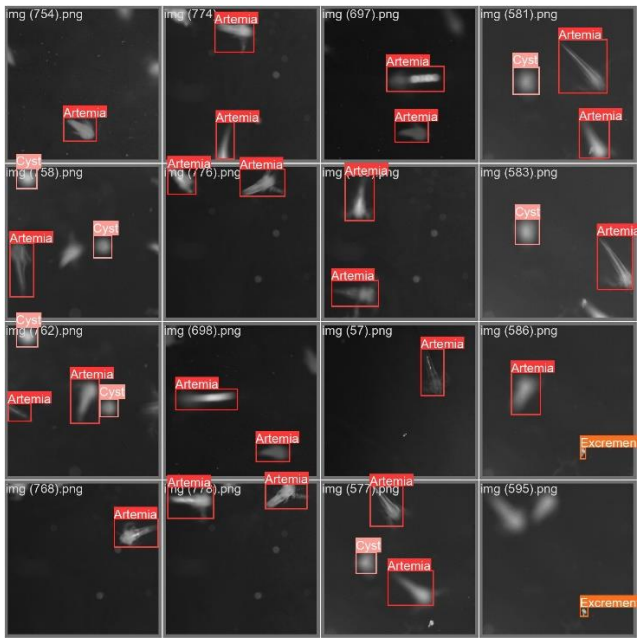

Figure 21. YOLOv8 pre-trained model training outputs

## C. Inferencing Outputs

To evaluate the object detection capabilities of YOLOv5 and YOLOv8 on real-world data, a total of 25 images were subjected to testing for the detection of Artemia, cyst, and excrement. These images were processed using both models, each trained on the labeled dataset.

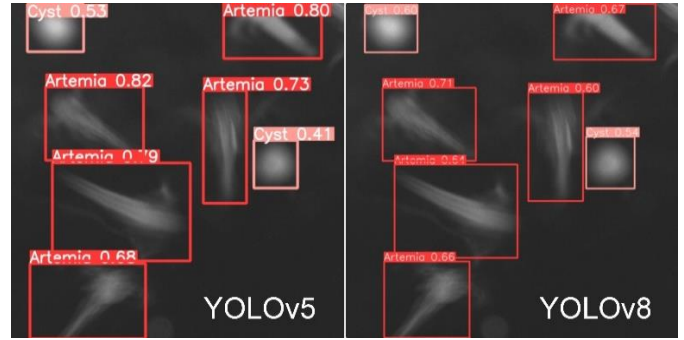

Figure 22. Inferencing result sample 1

What makes the inference results intriguing is the subtle performance differences observed across the two models:

YOLOv5's Artemia and Cyst Detection: In several cases (Figures 22 and 23), YOLOv5 outperformed YOLOv8 in the detection of Artemia and cyst. Its ability to accurately identify and classify these objects showcased its proficiency, particularly in scenarios where precision and accuracy were paramount.

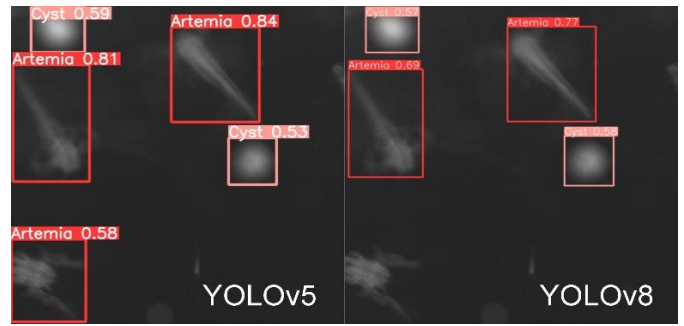

Figure 23. Inferencing result sample 2

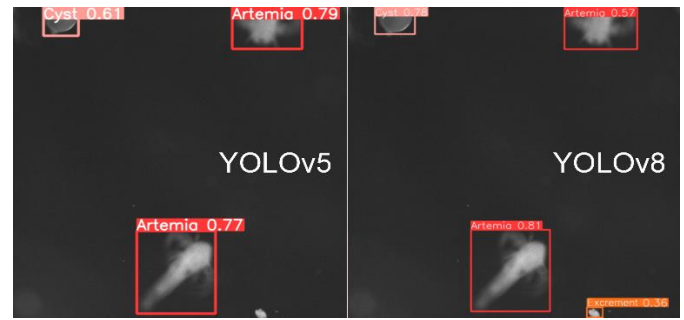

Figure 24. Inferencing result sample 3

Challenging Excrement Detection: However, it is noteworthy that YOLOv5 faced challenges when it came to the detection of excrement. In these instances (Figure 24), YOLOv5 exhibited limitations, as it struggled to identify and detect excrement accurately. This observation suggests that YOLOv5 may require further fine-tuning or specialized training for enhanced performance in this specific detection task.

One of the reasons causing this difference between YOLOv5 and YOLOv8 in detecting excrement can be the utilization of DFL (Distributional Focal Loss) in YOLOv8. It plays a crucial role in addressing the challenges associated with object detection. DFL functions are used for bounding box loss and binary cross-entropy for classification loss. These losses have been specifically designed to enhance smaller object detection, excrement in our case. It extends Focal Loss from discrete to continuous labels, that optimize and improve quality estimation and class prediction [8]. It enables YOLOv8 to provide a more accurate representation of the flexible distribution present in real data, thus reducing the risk of inconsistencies in detection results.

Another key advantage of DFL is its ability to handle class imbalance effectively [9]. In our dataset, there is a huge imbalance between the classes. DFL assigns higher weights to challenging examples, excrement in our case. This function allows the network to focus on learning the probabilities of values around the continuous locations of target bounding boxes [10]. This ensures that the model can adapt to arbitrary and flexible distributions, improving its ability to detect and classify objects accurately, even in challenging scenarios.

With these findings, the choice between YOLOv5 and YOLOv8 should be driven by the specific requirements and variations of the task. Further studies and more extensive training may help confirm these initial observations and guide the selection of the most suitable model for a given application.

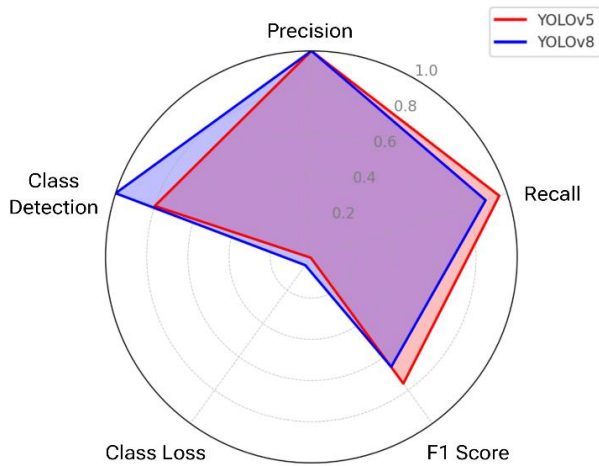

Figure 25. Radar chart comparing performance of YOLOv5 and YOLOv8

## V. CONCLUSION

The outcomes and results derived from the evaluation of YOLOv5 and YOLOv8 in object detection present an intriguing and subtle picture. The analysis of SSIM in our study further evaluates the context of object detection performance between YOLOv5 and YOLOv8. The high SSIM values observed across various image concentrations emphasize the importance of maintaining consistent image quality and structural integrity in training datasets. For example, the precision in detecting Artemia and cyst using YOLOv5 suggests that its performance benefits significantly from high-quality, structurally similar images. One the other hand, the slight overall decrease in SSIM values with increased concentration levels indicates the subtle challenges YOLOv8 faces in maintaining detection accuracy across varying image qualities and structural similarities. While both models exhibit strengths and capabilities, the findings suggest that YOLOv5 may excel over YOLOv8 in certain scenarios. However, it is equally apparent that the performance of these models can be context-dependent and class-specific.

One of the noteworthy observations from the evaluation is that YOLOv5 demonstrated superior performance in detecting *Artemia* and cyst in several instances, showcasing its potential in precision-driven detection tasks. Its agility and accuracy in these areas are promising, particularly for applications where accurate object recognition is paramount.

However, it is equally important to acknowledge that YOLOv5 exhibited limitations, particularly in the detection of less-represented classes, such as excrement. In cases where there are limited instances of a class within the labeled dataset, YOLOv5 appeared to struggle in accurately detecting those class objects.

On the other hand, YOLOv8 demonstrated robustness in detecting objects across a wider range of classes, even in scenarios with limited instances of a class. This observation implies that YOLOv8 may offer greater versatility and adaptability in certain detection tasks, even when training data is scarce for specific classes.

ACKNOWLEDGMENT

This work is supported by the National Science Foundation (award number: 2038484, year: 2020).